\documentclass[sigconf,nonacm]{acmart}
\usepackage{amssymb}
\usepackage{threeparttable}
\usepackage{multirow}
\usepackage{tikz}
\usepackage{pgfplotstable}
\usepackage{xcolor}
\usepackage{subcaption}
\usepackage{enumitem}
\usepackage{float}
\usepackage{stfloats}

\usepackage{makecell}

    \def\addlegendimage{\csname pgfplots@addlegendimage\endcsname}
\definecolor{acc}{HTML}{1A237E}
\definecolor{recall}{HTML}{B71C1C}
\definecolor{TrajCogn}{HTML}{1A237E}
\definecolor{prop1}{HTML}{B71C1C}
\definecolor{prop0.6}{HTML}{96C37D}
\definecolor{prop0.2}{HTML}{F3D266}
\definecolor{r50}{HTML}{1A237E}
\definecolor{r60}{HTML}{B71C1C}
\definecolor{r70}{HTML}{1B5E20}
\definecolor{r80}{HTML}{3E2723}

\AtBeginDocument{%
  \providecommand\BibTeX{{%
    \normalfont B\kern-0.5em{\scshape i\kern-0.25em b}\kern-0.8em\TeX}}}

\setcopyright{acmlicensed}
\copyrightyear{2018}
\acmYear{2018}
\acmDOI{XXXXXXX.XXXXXXX}

\acmISBN{978-1-4503-XXXX-X/18/06}
\newtheorem{definition}{Definition}

\newcommand{\vecbold}[1]{\boldsymbol{#1}}

\begin{document}

\title{TrajCogn: Leveraging LLMs for Cognizing Movement Patterns and Travel Purposes from Trajectories}

\author{Zeyu Zhou}
\authornote{Both authors contributed equally to this research.}
\email{zeyuzhou@bjtu.edu.cn}
\affiliation{%
  \institution{Beijing Jiaotong University}
  \city{Beijing}
  \country{China}
}

\author{Yan Lin}
\authornotemark[1]
\email{lyan@cs.aau.dk}
\affiliation{%
  \institution{Aalborg University}
  \city{Aalborg}
  \country{Denmark}
}

\author{Haomin Wen}
\email{haominwe@andrew.cmu.edu}
\affiliation{%
  \institution{Carnegie Mellon University}
  \city{Pittsburgh}
  \country{USA}
}

\author{Qisen Xu}
\author{Shengnan Guo}
\email{{23218133,guoshn}@bjtu.edu.cn}
\affiliation{%
  \institution{Beijing Jiaotong University}
  \city{Beijing}
  \country{China}
}

\author{Jilin Hu}
\email{jlhu@dase.ecnu.edu.cn}
\affiliation{%
  \institution{East China Normal University}
  \city{Shanghai}
  \country{China}
}

\author{Youfang Lin}
\author{Huaiyu Wan}
\authornote{Corresponding author.}
\email{{yflin,hywan}@bjtu.edu.cn}
\affiliation{%
  \institution{Beijing Jiaotong University}
  \city{Beijing}
  \country{China}
}

\renewcommand{\shortauthors}{Zhou and Lin, et al.}

\begin{abstract}
Spatio-temporal trajectories are crucial in various data mining tasks. It is important to develop a versatile trajectory learning method that performs different tasks with high accuracy. This involves effectively extracting two core aspects of information--movement patterns and travel purposes--from trajectories. However, this is challenging due to limitations in model capacity and the quality and scale of trajectory datasets. Meanwhile, large language models (LLMs) have shown great success in versatility by training on large-scale, high-quality datasets. Given the similarities between trajectories and sentences, there's potential to leverage LLMs to develop an effective trajectory learning method. However, standard LLMs are not designed to handle the unique spatio-temporal features of trajectories and cannot extract movement patterns and travel purposes.

To address these challenges, we propose a model called TrajCogn that effectively utilizes LLMs to model trajectories. TrajCogn leverages the strengths of LLMs to create a versatile trajectory learning approach while addressing the limitations of standard LLMs. First, TrajCogn incorporates a novel trajectory semantic embedder that enables LLMs to process spatio-temporal features and extract movement patterns and travel purposes. Second, TrajCogn introduces a new trajectory prompt that integrates these patterns and purposes into LLMs, allowing the model to adapt to various tasks. Extensive experiments on two real-world datasets and two representative tasks demonstrate that TrajCogn successfully achieves its design goals. Codes are available at \url{https://anonymous.4open.science/r/TrajCogn-5021}.
\end{abstract}

%    <concept>
%        <concept_id>10010147.10010257.10010293.10010294</concept_id>
%        <concept_desc>Computing methodologies~Neural networks</concept_desc>
%        <concept_significance>500</concept_significance
%        </concept>
%    <concept>
%        <concept_id>10010147.10010178</concept_id>
%        <concept_desc>Computing methodologies~Artificial intelligence</concept_desc>
%        <concept_significance>300</concept_significance>
%        </concept>
%  </ccs2012>

% \ccsdesc[500]{Computing methodologies~Neural networks}
% \ccsdesc[300]{Computing methodologies~Artificial intelligence}

% \keywords{Spatio-temporal data mining,trajectory data mining,large language models}


\maketitle

\section{Introduction}
A spatio-temporal (ST) trajectory is a sequence of timestamped locations, represented as $\mathcal T=\langle (l_1,t_1), (l_2,t_2), \dots, (l_n,t_n) \rangle$. It tracks the movements of an individual or object in a geographical space. With the widespread use of mobile phones, car navigation systems, location-based services, and online map services, ST trajectories are being recorded and collected from various sources~\cite{geolife}. They enable a wide range of spatio-temporal data mining tasks and applications, including trajectory prediction~\cite{deepmove, hst-lstm}, anomaly detection~\cite{gcm, gmvsae}, trajectory similarity measurement~\cite{neutraj, st2vec, t2vec, t3s}, and trajectory-user linking~\cite{tul, deeptul}.

To enhance the use of ST trajectories in tasks and applications, it is essential to develop a trajectory learning method that 1) effectively captures the information embedded in the trajectory, specifically, movement patterns that describe how the individual or object moves from one location to another and travel purposes that indicate the underlying reason or motivation for the movement; and 2) accurately performs a variety of downstream tasks, reducing the need for designing a separate method for each task and application. Existing efforts mostly adhere to the self-supervised representation learning approach~\cite{semi,bert}, which builds a trajectory learning model that maps a trajectory into its embedding vector and trains the model from scratch~\cite{lightpath,trembr,start}. However, considering the complexity of the information embedded in trajectories and the difficulty in creating a versatile learning model capable of performing different tasks, the effectiveness of these existing models is limited by their capacities and the size and quality of available trajectory datasets.

\begin{figure}[t]
    \centering
    \includegraphics[width=1.0\linewidth]{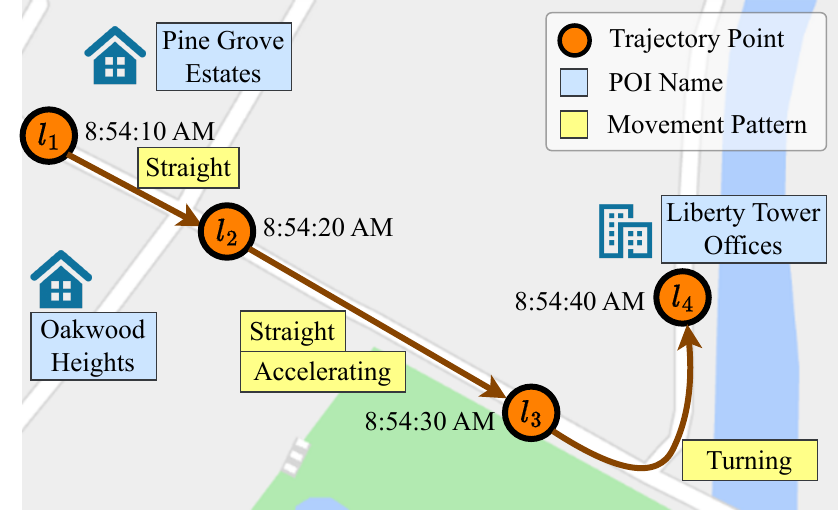}
    \caption{A trajectory of commuting to work.}
    \label{fig:traj-example}
\end{figure}

On the other hand, versatile models have been highly successful in the domain of natural language processing (NLP), showcasing promising results on various downstream tasks~\cite{gpt2, bert, t5, glm}. These models, often referred to as large language models (LLMs), benefit mainly from their large capacity, abundant large-scale corpus datasets, and well-thought-out prompt engineering~\cite{gpt3}. Given the similarities between trajectories and sentences in NLP, there is significant potential in building a more effective trajectory learning model by leveraging LLMs. Trajectory points exhibit spatio-temporal correlations similar to the contextual correlations between words in sentences. Additionally, movement patterns in trajectories, such as turning and acceleration, can be considered akin to the semantics of words. Furthermore, the travel purpose of trajectories, such as leisure activities or commuting, can be seen as similar to the semantics of sentences. Despite this potential, there are two challenges in adapting LLMs to model trajectories.

\textbf{First, LLMs are incapable of processing the spatio-temporal features in trajectories.}
LLMs are designed to handle sequences of discrete word tokens as input. However, trajectories consist of both continuous and discrete spatio-temporal features, such as GPS coordinates, timestamps, and road segments. It is challenging to process these features in a way that LLMs can understand and extract information from.

\textbf{Second, LLMs are unable to extract the movement patterns and travel purposes directly from trajectories.}
The movement patterns in trajectories are represented by the changes between features of trajectory points. Take Figure~\ref{fig:traj-example} as an example, where the moving object goes straight from points $(l_1,t_1)$ to $(l_3,t_3)$, accelerates between points $(l_2,t_2)$ and $(l_3,t_3)$, and turns left between points $(l_3,t_3)$ and $(l_4,t_4)$. These patterns can be derived from changes in coordinates, timestamps, and velocities. Moreover, the travel purpose of a trajectory is closely linked to its origin and destination (OD). As shown in Figure~\ref{fig:traj-example}, the trajectory originates near several residential buildings and concludes near an office building, indicating that the travel purpose of this trajectory is commuting. However, LLMs primarily focus on modeling the semantic meaning of words in a sentence. They lack the necessary design to effectively extract movement patterns from spatio-temporal features, or to model travel purposes from the functionalities of locations near a trajectory's OD.

To address these challenges and effectively leverage LLMs to construct a versatile trajectory learning model, we propose a novel approach named \textit{\underline{Traj}ectory \underline{Cogn}ition} (\textbf{TrajCogn}).
TrajCogn employs a trajectory prompt to integrate movement patterns and travel purposes from trajectories. Additionally, by implementing the task-p-tuning mechanism in the prompt, TrajCogn can adapt to various downstream tasks and generate accurate predictions. TrajCogn also encompasses a trajectory semantic embedder to enable LLMs to process the spatio-temporal features in trajectories and effectively extract movement patterns and travel purposes. To enhance the training of TrajCogn, we implement a cross-reconstruction pretext task based on self-supervised reconstruction. This improves the model's ability to learn from trajectory data.
Our contributions are summarized as follows:
\begin{itemize}[leftmargin=*]
    \item We propose TrajCogn, a model that effectively migrates LLMs to cognize movement patterns and travel purposes from trajectories. By taking advantage of the adaptability of LLMs, TrajCogn accurately performs different downstream tasks, mitigating the limitation of small-scale trajectory datasets.
    \item We introduce a novel trajectory prompt that integrates the two essential aspects of information in trajectories, namely movement patterns and travel semantics, into LLMs. This prompt also enables the model to effectively adapt to various downstream tasks.
    \item We propose a novel trajectory semantic embedder that enables LLMs to process the spatio-temporal features of trajectories. This embedder ensures that LLMs can effectively extract movement patterns and travel semantics in an explainable manner.
    \item We conduct extensive experiments on two real-world trajectory datasets to evaluate the proposed model with a variety of experimental settings. The results showcase that TrajCogn is a versatile trajectory learning model that demonstrates strong performance across different tasks.
\end{itemize}

\section{Related Works}
\textbf{Trajectory Learning Models} aim to extract information from trajectories and perform various related tasks. Compared to task-specific prediction models~\cite{deepmove,neutraj,st2vec,deepeta,hiereta,DBLP:conf/kdd/LiFWSYL18}, which are end-to-end trained for one specific task, trajectory learning models are versatile and useful in modern intelligent transportation applications that usually involve multiple tasks.

Most existing efforts adhere to the self-supervised learning approach. Earlier research commonly used RNNs to reconstruct discrete locations~\cite{t2vec, gmvsae, trembr} or continuous movement features~\cite{traj2vec} of trajectories based on auto-encoding~\cite{ae} and variational auto-encoders~\cite{vae}. Additionally, methods like CTLE~\cite{ctle} and Toast~\cite{toast}, based on transformers~\cite{transformer} and Masked Language Model (MLM) tasks~\cite{bert}, treat trajectory points as tokens in a sentence. Furthermore, contrastive learning methods such as PIM~\cite{pim}, TrajCL~\cite{trajcl}, and MMTEC~\cite{mmtec} implicitly model the travel purpose of a trajectory. More recently, methods combining multiple approaches have been developed. START~\cite{start} leverages both MLM tasks and SimCLR~\cite{simclr}, while LightPath~\cite{lightpath} incorporates a reconstruction task and a contrastive-style rational reasoning task.

Since these methods are self-supervised and trained from scratch, their performance heavily relies on the size and quality of the training datasets, which often have limitations. Despite the achievements of existing methods, further efforts are needed to enhance the performance of trajectory learning models.

\textbf{Cross-domain Application of LLMs}.
The versatility and superior performance of large language models (LLMs) in the NLP domain have led to efforts to adopt LLMs in other fields to enhance performance. In time series analysis, GPT4TS~\cite{GPT4TS} uses LLMs by freezing the self-attention feed-forward layers. Time-LLM~\cite{timellm} introduces a reprogramming framework. For visual encoding tasks, LM4VisualEncoding~\cite{lm4visual} incorporates a frozen transformer block from an LLM as a general-purpose visual encoder layer. RLMRec~\cite{rlmrec} integrates the semantic space of LLMs with collaborative relational signals using an alignment framework.

Although these studies provide valuable insights, their methods cannot be directly applied to trajectory learning. Trajectory data has unique spatio-temporal features that require tailored approaches and considerations.

\section{Preliminaries}
\subsection{Definition}
\begin{definition}[Road Network]
    A road network is represented as a directed graph $\mathcal{G} = (\mathcal{V}, \mathcal{E})$. 
    $\mathcal{V}$ is a set of $\vert\mathcal{V}\vert$ vertices, and each vertex $v_i\in \mathcal{V}$ represents an intersection between road segments or the end of a segment. 
    $\mathcal{E}$ is a set of $\vert \mathcal{E}\vert$ segments, where each segment $s_i\in\mathcal{E}$ represents a road segment linking two vertices. 
\end{definition}

\begin{definition}[Trajectory]
    A trajectory $\mathcal{T}$ is a sequence of timestamped locations, represented as $\mathcal{T} = \langle (l_1,t_1), (l_2,t_2), \cdots, (l_n,t_n) \rangle$. Here, each location $l_i$ is represented by its latitude and longitude coordinates, i.e., $l_i = (l^\mathrm{lat}_i, l^\mathrm{lng}_i)$. The timestamp $t_i$ indicates when $l_i$ is visited. To simplify, we denote the $i$-th trajectory point $(l_i, t_i)$ as $\tau_i$. 
\end{definition}

\begin{definition}[Point of Interest, POI]
   A POI is a particular location that individuals may find valuable or intriguing. It is denoted as $p=(l, n, a)$, where $l$ represents its coordinates, $n$ indicates its name, and $a$ refers to its address.
\end{definition}

\subsection{Problem Statement}
\noindent \textit{Trajectory Learning.} 
The objective is to develop a trajectory learning model $f_{\Theta}$ with a set of learnable parameters $\Theta$. This model takes a trajectory $\mathcal{T}$ as input and extracts information from it. Subsequently, this model can adapt to various downstream tasks by accurately predicting the required outputs $y$ for the task at hand, denoted as $\hat{y} = f_\Theta(\mathcal{T})$. For example, in travel time estimation, $y$ and $\hat{y}$ represent the ground truth and the estimated travel time, respectively.

\subsection{Pre-trained Language Model}
    In this work, a Large Language Model (LLM) refers to a \linebreak 
    Transformer-based language model pre-trained on corpus datasets. It consists of four essential functions. Formally,
    \begin{equation}
        \label{equ:LLM}
        \mathrm{LLM} = \mathrm{LMHead}\circ\mathrm{TransBlk}\circ\mathrm{WTE}\circ\mathrm{Tok}(\cdot),
    \end{equation}
    where $\circ$ represents the composition of functions. Specifically, a LLM consists of a tokenizer ($\mathrm{Tok}$) to break down text into discrete tokens, a word token embedding layer ($\mathrm{WTE}$) that converts the tokens into numerical vectors to capture their linguistic features, a transformer block ($\mathrm{TransBlk}$) that further processes the vectors to capture their contextual relationships, and a prediction head ($\mathrm{LMHead}$) that is respondible for making specific predictions, such as generating the next word in a sequence. In a LLM, the dimension of the word token embedding is denoted as $d$.

\section{Methodology}

\subsection{Overview}

\begin{figure*}[t]
    \centering
    \includegraphics[width=1.0\linewidth]{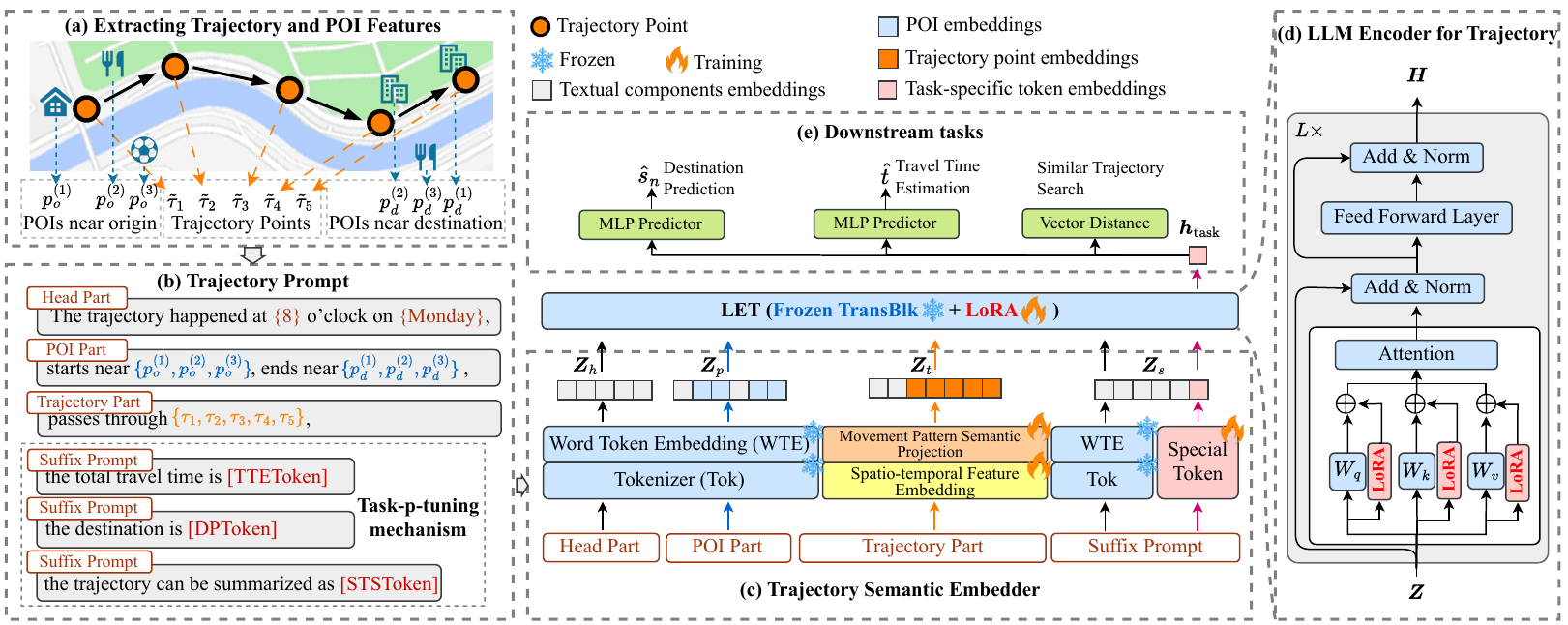}
    \caption{Overall framework of TrajCogn.}
    \label{fig:overview}
\end{figure*}

Figure~\ref{fig:overview} shows the overall framework of TrajCogn. It is implemented in the following four steps:
\begin{enumerate}[leftmargin=*]
    \item \textbf{Trajectory and POI Feature Extraction}: Given a trajectory $\mathcal{T}$, we perform map-matching and calculate high-order features such as velocity, acceleration, and direction to expand its features, denoted as $\widetilde{\mathcal{T}}$. We also extract the address and name features of POIs near the trajectory’s origin and destination.
    \item \textbf{Trajectory Prompt Construction}: We integrate the extracted features into one sequence, called the trajectory prompt. This prompt also includes a task-p-tuning mechanism-based suffix to enable adaptation to various tasks.
    \item \textbf{Trajectory Prompt Embedding}: We map the trajectory prompt into a sequence of $d$-dimensional embeddings with a trajectory semantic embedder. This embedder is designed to enable LLMs to process spatio-temporal features and effectively extract movement patterns and travel purposes with explainability.
    \item \textbf{Model Training and Task Adaptation}: We process the embedding sequence with a \textit{\underline{L}LM \underline{E}ncoder for \underline{T}rajectory} (LET). The last point of the output sequence of LET is used for performing downstream tasks. The learnable parameters in the model are refined by integrating a cross-reconstruction pretext task and further optimized with a dedicated objective function for each specific downstream task.
\end{enumerate}

The following sections provide a detailed explanation of the steps in TrajCogn.

\subsection{Trajectory Prompt}
As illustrated in Figure~\ref{fig:traj-example}, movement patterns in a trajectory can be represented by positions on the road network and variations in spatio-temporal features. Travel purposes can be inferred from the functionalities of locations near the OD points, and the address and name features of a POI indicate its functionalities.

To incorporate the movement patterns and travel purposes of a trajectory, we first extract spatio-temporal and POI features from the trajectory, as shown in Figure~\ref{fig:overview}(a). To integrate these features into LLMs, we introduce a \textit{Trajectory Prompt}, as illustrated in Figure~\ref{fig:overview}(b). This prompt fuses natural language and the extracted features into a sequence. Furthermore, to adapt the model to different downstream tasks, we introduce a task-p-tuning mechanism, which provides a specific suffix for each task.

\subsubsection{Trajectory and POI Feature Extraction}
\label{subsubsec:extract-trajectory-features} 
Given a trajectory $\mathcal T=\langle (l_1,t_1), (l_2,t_2), \dots, (l_n,t_n) \rangle$ and the road network $\mathcal{G}$, we utilize the Leuven Map Matching (LMM) algorithm~\cite{leuven} to map each trajectory point $\tau_i$ onto the road network. This mapping is denoted as $\text{LMM}(\tau_i, \mathcal G)=(l_i,s_i,t_i)$, where $s_i$ represents the road segment on which $l_i$ is located. We also calculate the velocity $v_i$, acceleration $a_i$, and direction $\theta_i$ of each trajectory point $\tau_i$ according to the difference between the features of $\tau_i$ and $\tau_{i+1}$. Next, we gather the trajectory point $\tilde{\tau_i} = (l_i,s_i,t_i, v_i, a_i, \theta_i)$ with extracted spatio-temporal features. We set the velocity and acceleration of the last point $\tilde{\tau_n}$ to 0. Finally, we obtain the trajectory $\widetilde{\mathcal T}=\langle \tilde\tau_1, \tilde\tau_2, \dots, \tilde\tau_n \rangle$ with extracted features.

To extract POI features, we begin by identifying the origin $l_1$ and destination $l_n$ of trajectory $\mathcal{T}$. Using the Ball Tree algorithm~\cite{balltree}, we retrieve the closest $N_\text{POI}$ POIs to $l_1$. The set of retrieved POIs is denoted as $\mathcal{P}_O$, where $\mathcal{P}_O = \{ p_o^{(1)}, \dots, p_o^{(N_{\text{POI}})} \}$, and the POIs in $\mathcal{P}_O$ are arranged in ascending order based on their distance from the origin. Similarly, we retrieve the set of POIs around $l_n$ as $\mathcal{P}_D = \{ p_d^{(1)}, \dots, p_d^{(N_\text{POI})} \}$.
For each POI $p\in\mathcal{P}_O\cup\mathcal{P}_D$, we extract its address $p.a$ and name $p.n$ features, both represented as lists of words.

\subsubsection{Trajectory Prompt Construction}
\label{subsec:trajectory-prompt}
The trajectory prompt is composed of four parts, defined as follows:
    \begin{enumerate}[leftmargin=*]
        \item $\langle$Head Part$\rangle$: \textit{"The trajectory happened on { \{day-in-week\}} at \{hour\} o'clock, "}
        \item $\langle$POI Part$\rangle$: \textit{"starts near: \{$p_o^{(1)}, p_o^{(2)} , \dots, p_o^{(N_\mathrm{POI})}$\}, ends near: \{$p_d^{(1)}, \\p_d^{(2)} , \dots, p_d^{(N_\mathrm{POI})}$\},"}
        \item $\langle$Trajectory Part$\rangle$: \textit{"passes through \{$\tilde\tau_1, \tilde\tau_2, \dots, \tilde\tau_n$\}."}
        \item $\langle$Suffix Prompt$\rangle$
    \end{enumerate}

The $\langle$Head Part$\rangle$ enriches the input context and guides the LLM in analyzing trajectories. The $\langle$POI Part$\rangle$ provides information about the addresses and names of the POIs around the OD points, allowing the LLM to infer travel purposes. The $\langle$Trajectory Part$\rangle$ comprises the extracted features of the trajectory points, enabling the model to extract movement patterns. The placeholders \{\} are filled with trajectory-specific features and information.

The $\langle$Suffix Prompt$\rangle$ is constructed using the proposed task-p-tuning mechanism, enabling the model to perform different downstream tasks. It is a hybrid of hard and soft components~\cite{ptr}. The hard component consists of words that signify the particular task. The soft component $\text{[Token]}$ is a task-specific token with a learnable embedding vector. For example, in the case of travel time estimation (TTE), the suffix prompt would be "the total travel time is [TTEToken]." Similarly, for destination prediction (DP), the suffix prompt would be "the destination is [DPToken]."

\subsection{Trajectory Semantic Embedder}
In order to equip LLMs with the ability to process the spatio-temporal features in the trajectory prompt, we propose the \textit{Trajectory Semantic Embedder}, demonstrated in Figure~\ref{fig:overview}(c). 

\subsubsection{Spatio-temporal Feature Embedding}
We embed the spatio-temporal features in the constructed trajectory prompt into a $d$-dimensional embedding space. For the discrete road segment $s_i$, we use an index-fetching embedding module $\vecbold{E}_{\mathcal{E}} \in \mathbb{R}^{\vert \mathcal{E} \vert \times d}$. The embedding vector for road segment $s_i$ is represented as $\vecbold{E}_{\mathcal{E}}(s_i)$. Similarly, for the timestamp $t_i$, we use two index-fetching embedding modules: $\vecbold{E}_{\mathrm{dw}} \in \mathbb{R}^{7 \times d}$ and $\vecbold{E}_{\mathrm{h}} \in \mathbb{R}^{24 \times d}$ to embed the cyclic time features, namely day-in-week and hour, as $\vecbold{E}_{\mathrm{dw}}(t_i)$ and $\vecbold{E}_{\mathrm{h}}(t_i)$ respectively.

To facilitate the modeling of movement patterns from variations of continuous features, we take inspiration from previous studies~\cite{deepeta, trajformer} and employ a one-dimensional convolution for embedding continuous features. 
Given the continuous features of the $i$-th trajectory point, denoted as $\tau_i^\text{con} = (l^{\mathrm{lat}}_i, l^{\mathrm{lat}}_i, v_i, a_i, \theta_i, t_i)$, the convolution on this point is formulated as follows:
\begin{equation}
    \label{equ:conv}
    \vecbold{E}_\mathrm{con}(i) = \mathrm{Conv1D}(\tau_{i-\lfloor\frac{k}{2}\rfloor:i+\lfloor\frac{k}{2}\rfloor}^\text{con}),
\end{equation}
where $k$ is a hyper-parameter, denoting the kernel size, and $\vecbold{E}_{\mathrm{con}}(i)\in \mathbb{R}^{d}$ represents the continuous embedding vector of $\tau_i$.

Finally, the embedding vector $\vecbold{e}_i$ of the $i$-th trajectory point $\tau_i$ is derived as follows:
\begin{equation}
    \label{equ:fea-embedding}
    \vecbold{e}_i = \vecbold{E}_{\mathrm{con}}(i) + \vecbold{E}_{\mathcal{E}}(s_i) + \vecbold{E}_{\mathrm{dw}}(t_i) + \vecbold{E}_{\mathrm{h}}(t_i)
\end{equation}

\subsubsection{Movement Pattern Semantic Projection}
\begin{figure}[t]
    \centering
    \includegraphics[width=0.9\linewidth]{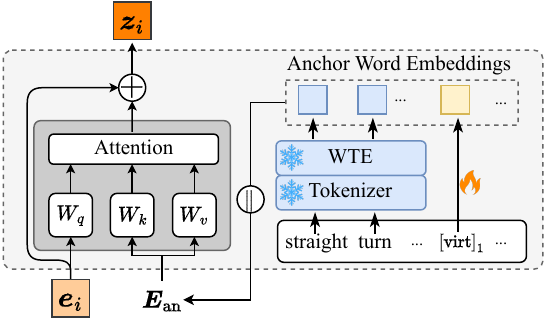}
    \caption{Movement pattern semantic projection.}
    \label{fig:psp}
\end{figure}

To enhance the model's ability to understand the semantics of movement patterns and improve its interpretability, we project each embedding vector $\vecbold{e}_i$ onto a semantic-rich textual space, as shown in Figure~\ref{fig:psp}.

The textual space is defined by a set of words that we choose to describe the movement patterns. Specifically, we establish a set of words $\mathcal{M}$, with its content listed in Table~\ref{tab:word-list}. 
For words in $\mathcal{M}$, we obtain their embedding vectors using LLM components $\mathrm{WTE}\circ\mathrm{Tok}$ introduced in Equation~\ref{equ:LLM}.

\begin{table}[h]
    \centering
    \caption{Words describing movement patterns.}
        \begin{tabular}{cm{5cm}}
        \toprule
        Categories & \makebox[5cm]{Words} \\
        \hline 
        Driving Behaviors & \small{straight, turn, u-turn, brake, accelerate, decelerate, stop, overtake, zigzag, swerve, detour, slide, cruise, glide, cautious, reckless, leisurely} \\
        \hline
        Traveling Dynamics & \small{ steady, smooth, rough, constant, dynamic, fast, slow, rapid, rushed, erratic, agile, stationary, sluggish} \\
        \bottomrule
    \end{tabular}

    \label{tab:word-list}
\end{table}

Furthermore, we introduce a set $\mathcal{A}$ of virtual words:
\begin{equation}
    \mathcal{A} = \langle \text{"[virt]$_1$"}, \text{"[virt]$_2$"}, \text{"[virt]$_{N_A}$"} \rangle,
\end{equation}
where their word embeddings are initialized randomly and trained end-to-end, and $N_A$ is the number of virtual words. The words in $\mathcal{M} \cup \mathcal{A}$ are termed anchor words. 
We concatenate the embeddings of all anchor words, denoted as $\vecbold{E}_\mathrm{an}\in\mathbb{R}^{(\vert\mathcal{M}\vert + N_A) \times d}$.

To project an embedding vector $\vecbold{e}_i$ onto the space defined by $\vecbold{E}_\mathrm{an}$, we employ a dot-product multi-head attention~\cite{transformer} with $N_H$ attention heads. 
The attention is calculated using $\vecbold{e}_i$ as query and $\vecbold{E}_{\mathrm{an}}$ as key and value:
\begin{equation}
    \label{finalembedding}
    \widetilde{\vecbold{z}}_i = \mathrm{Attention}(\vecbold{e}_i, \vecbold{E}_\mathrm{an}, \vecbold{E}_\mathrm{an})
\end{equation}
The final embedding vector $\vecbold{z}_i$ for each trajectory point is then obtained with a residual connection:
\begin{equation}
    \label{equ:zi}
    \vecbold{z}_i = \mathrm{MLP}(\widetilde{\vecbold{z}}_i) + \vecbold{e}_i,
\end{equation}
where $\mathrm{MLP}$ represents a two-layer fully connected network. Finally, the embedding sequence of $\widetilde{\mathcal{T}}$ is denoted as $\vecbold{Z}_\mathcal T = \langle \vecbold{z}_1, \vecbold{z}_2, \dots, \vecbold{z}_n \rangle$.

\subsubsection{POI Feature Embedding}
The travel purpose can be determined by analyzing the functionalities of POIs around the OD points, as shown in Figure~\ref{fig:traj-example}. To model the functionalities of POIs, we fetch their embeddings based on their address and name features.

Specifically, in the case of a POI $p$ is either $p_o^{(1)}$ or $p_d^{(1)}$ , which are the closest POIs to the origin or destination, we obtain its embedding as follows:
\begin{equation}
    \vecbold{E}_\text{Tok}(p) = \mathrm{WTE}\circ\mathrm{Tok}(p.a \Vert p.n),
\end{equation}
where $p.a$ and $p.n$ are the address and name of $p$, consists of list of words. $\Vert$ denotes list concatenation.
For the remaining POIs, we solely utilize their names to obtain their embeddings as $\vecbold{E}_\text{Tok}(p) = \mathrm{WTE}\circ\mathrm{Tok}(p.n)$.

\subsubsection{Sequence of Trajectory Prompt Embeddings}
After obtaining embeddings of spatio-temporal and POI features, the remaining textual components in the trajectory prompt are embedded using $\mathrm{WTE\circ Tok}$. Then, we concatenate the embeddings into a sequence in the same order as their raw features appear in the prompt.
For example, the embeddings of the trajectory part are obtained as follows:
\begin{equation}
    \vecbold{Z}_t = \vecbold{E}_\mathrm{Tok}(\text{"passes through"}) \Vert \vecbold{Z}_\mathcal T
\end{equation}

The embeddings of the $\langle$Head Part$\rangle$, $\langle$POI Part$\rangle$, and $\langle$Suffix Prompt$\rangle$ are denoted as $\vecbold{Z}_h$, $\vecbold{Z}_p$, and $\vecbold{Z}_s$, respectively. 
Finally, the sequence of trajectory prompt embeddings is gathered as follows:
\begin{equation}
    \label{equ:zd}    
    \vecbold{Z} = \vecbold{Z}_h \Vert \vecbold{Z}_p \Vert \vecbold{Z}_t \Vert \vecbold{Z}_s
\end{equation}

\subsection{LLM Encoder for Trajectory}
We use the transformer block $\mathrm{TransBlk}$ from an LLM as the backbone for the proposed \textit{LLM Encoder for Trajectory} (\textbf{LET}). To better adapt the pre-trained $\mathrm{TransBlk}$ to trajectory learning, we employ the Low Rank Adaptation (LoRA) algorithm~\cite{lora}, adding extra parameters to $\mathrm{TransBlk}$.

\subsubsection{Construction of LET}
As illustrated in Figure~\ref{fig:overview}(d), all parameters in the $\mathrm{TransBlk}$ are kept fixed, while we introduce a new learnable parameter matrix $\Delta \vecbold{W}_{*}$ of the same size for each of $\vecbold{W}_q, \vecbold{W}_k, \vecbold{W}_v$ in every self-attention block of $\mathrm{TransBlk}$. 
Each $\Delta \vecbold{W}$ is a low-rank matrix that can be written as the product of two low-rank matrices, i.e., $\Delta \vecbold{W} =  \vecbold{BA},  \vecbold{B}\in \mathbb{R}^{d\times r}, A\in \mathbb{R}^{r\times d}$, where $r$ is a hyper-parameter, denoting the rank of LoRA with $r \ll d$. 
The modified query matrix in each self-attention block of $\mathrm{TransBlk}$ is presented as $ \vecbold{Q} = (\vecbold{W}_q + \Delta \vecbold{W}_q) \vecbold{H} = \vecbold{W}_q \vecbold{H} +  \vecbold{B}_q \vecbold{A}_q \vecbold{H}$, where $ \vecbold{H}$ represents a hidden state of a model layer. The same modification is applied to the key and query matrices. 
Next, the proposed LET can be expressed as follows:
\begin{equation}
    \mathrm{LET} = \mathrm{LoRA}(\mathrm{TransBlk})
\end{equation}

LET takes the embedding sequence $ \vecbold{Z}$ from Equation~\ref{equ:zd} as input, and outputs a sequence of hidden vectors $ \vecbold{H}\in \mathbb{R}^{L\times d}$, where $L$ represents the length of $\vecbold{Z}$.
\begin{equation}
    \vecbold{H} = \mathrm{LET}( \vecbold{Z})
\end{equation}

\subsubsection{Adaptation to Downstream Tasks}
\label{subsubsec:downstream}
LET adapts to different downstream tasks using the task-p-tuning mechanism described in Section~\ref{subsec:trajectory-prompt}.
Specifically, the hidden vector corresponding to the task-specific token, i.e., the $L$-th hidden vector $\vecbold{h}_\mathrm{task} \in \mathbb{R}^{d}$ in $\vecbold{H}$, can be utilized to perform downstream tasks.

In this study, we present \textit{Travel Time Estimation} (\textbf{TTE}), \textit{Destination Prediction} (\textbf{DP}), and \textit{Similar Trajectory Search} (\textbf{STS}) tasks for evaluation, as shown in Figure~\ref{fig:overview}(e).

The TTE task aims to estimate the travel time of a trajectory given its spatial features and departure time, without using time-related features including timestamp, velocity, and acceleration. For this task, a prediction head is built using a two-layer fully connected network to obtain the prediction as follows:
\begin{equation}
    \hat{y}_{\mathrm{TTE}} = \mathrm{MLP}_\mathrm{TTE}(\vecbold{h}_\mathrm{task})
\end{equation}

The DP task aims to predict the road segment where the destination of a trajectory is located, given the trajectory excluding its last 5 points. To prevent data leakage, the trajectory prompt does not include any POIs near the destination while performing this task.
For this task, a prediction head is built with a two-layer fully connected network, where the output dimension corresponds to the total number of segments $\vert \mathcal{E} \vert$:
\begin{equation}
    \hat{y}_\mathrm{DP} = \mathrm{argmax}_{s}(\hat{ \vecbold{p}}), \hat{ \vecbold{p}} = \mathrm{Softmax}(\mathrm{MLP}( \vecbold{h}_\mathrm{task}))
\end{equation}

The STS task aims to find the most similar trajectory from a set of candidates given a query. We use cosine similarity on $\vecbold{h}_\mathrm{task}$ to determine the similarity between trajectories. Since most datasets do not provide ground truth for this task, we construct the ground truth following the method introduced in Appendix~\ref{sec:sts-gt-construction}.

\subsection{Model Training}
\label{subsec:training}
We propose a cross-reconstruction pretext task to train the learnable parameters in the model, helping it adapt to trajectories. Before performing a specific task, the model can be further fine-tuned with supervision for that task.

\subsubsection{Cross-reconstrution Pretext Task}
\label{subsubsec:pretext}
The proposed pretext task involves reconstructing each trajectory point given $\langle$Head Part$\rangle$ and $\langle$POI Part$\rangle$, and reconstructing each POI given $\langle$Head Part$\rangle$ and $\langle$Trajectory Part$\rangle$.

\begin{figure}[t]
    \centering
    \includegraphics[width=\linewidth]{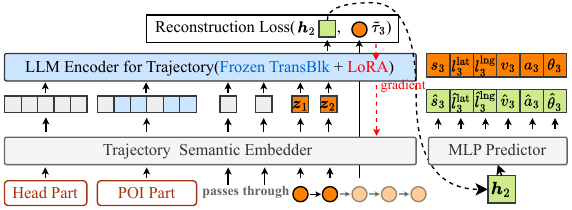}
    \caption{Reconstruction of trajectory points in cross-reconstruction pretext task.}
    \label{fig:cross-reconstruction}
\end{figure}

Firstly, we autoregressively reconstruct the trajectory point features, as shown in Figure~\ref{fig:cross-reconstruction}. Given a trajectory $\mathcal{T}$, this reconstruction consists of $\vert\mathcal{T}\vert$ steps. 
In the $i$-th step, LET receives the embeddings of trajectory prompt composed of $\langle$Head Part$\rangle$, $\langle$POI Part$\rangle$, and $\langle$Trajectory Part$\rangle$ with the first $i-1$ trajectory points:
\begin{equation}
    \label{equ:ar}
        \vecbold{H}_{\text{traj}, i-1} = \mathrm{LET}(\vecbold{Z}_h \Vert \vecbold{Z}_p \Vert \vecbold{Z}_{t,:i-1})
\end{equation}
Afterwards, we obtain predicted features by applying prediction heads on the last vector in $\vecbold{H}_{\text{traj}, i-1}$. All prediction heads are implemented with two-layer fully connected networks. 
The loss $\mathcal{L}_\text{traj}$ for trajectory reconstruction is then calculated by summing the cross-entropy loss of the predicted segments and the MSE loss of the predicted continuous features.

Next, we proceed with the reconstruction of the POI features. 
Similar to Equation~\ref{equ:ar}, in the $i$-th step, LET receives the embeddings of the trajectory prompt composed of $\langle$Head Part$\rangle$, $\langle$Trajectory Part$\rangle$, and $\langle$POI Part$\rangle$ with the first $i-1$ POIs:
\begin{equation}
    \label{equ:poi-recon}
    \vecbold{H}_{\text{POI},i-1}=\mathrm{LET}(\vecbold{Z}_h \Vert \vecbold{Z}_t \Vert \vecbold{Z}_{p,:i-1})
\end{equation}
Then, we obtain the predicted POI features by applying the $\mathrm{LMHead}$ component of LLMs on the last vector in $\vecbold{H}_{\text{POI},i-1}$.
The loss $\mathcal{L}_\text{POI}$ for POI reconstruction is the cross-entropy loss of the predicted POI features.

Finally, the loss function of the pretext task is represented as:
\begin{equation}
    \label{equ:pretext}
    \mathcal{L}_\text{pre} = \mathcal{L}_\mathrm{traj} + \mathcal{L}_\mathrm{POI}
\end{equation}

To improve the training efficiency, we utilize the teacher-forcing mode~\cite{teacher-forcing} to parallelize the reconstruction process.

% A specific projector is trained for each downstream task, and it is a MLP projector. 
% When undertaking specific downstream tasks such as TTE, we fine-tune all learnable parameters in the our proposed method, using the corresponding loss specific to that task such as $\mathcal{L}_{\text{TTE}}$. The detailed implementations are shown next.

\subsubsection{Task-specific Fine-tuning}
\label{subsubsec:downstream-train}
When performing a specific task, the proposed model can be fine-tuned with the task's supervision to further improve prediction accuracy.

For the TTE task, the loss function is defined with mean square error (MSE) loss:
\begin{equation}
    \mathcal{L}_{\mathrm{TTE}} = \frac{1}{2}\Vert \hat{y}_{\mathrm{TTE}} - y_{\mathrm{TTE}} \Vert_2^2
\end{equation}

For the DP task, the loss function is defined with the cross-entropy loss:
\begin{equation}
    \mathcal{L}_\mathrm{DP} = -\log\hat{\vecbold{p}}(s_d),
\end{equation}
where $s_d$ represents the label of the destination segment, and $\hat{\vecbold{p}}(s_d)$ denotes the $s_d$-th value of the predicted probability score $\hat{\vecbold{p}}$. 

For the STS task, no fine-tuning is involved. We directly use the hidden state $\vecbold{h}_\mathrm{task}$ from the cross-reconstruction pretext task.

\section{Experiments}
To evaluate the proposed method's effectiveness on trajectory learning, we conduct experiments on two real-world datasets and compare its performance against several state-of-the-art baselines.

\subsection{Datasets}
In our experiments, we use two real-world datasets called Chengdu and Xi'an. These datasets were released by Didi\footnote{https://gaia.didichuxing.com/} and consist of GPS trajectories recorded by taxis in Chengdu and Xi'an, China. Trajectories shorter than 6 points are excluded from our study. We fetch road networks covering the two datasets from OpenStreetMap\footnote{https://www.openstreetmap.org/} to map-match trajectories. An overview of the dataset statistics is shown in Table~\ref{tab:dataset}.

\begin{table}[h]
    \centering
    \caption{Dataset statistics.}
    \begin{tabular}{ccc}
    \toprule
    Dataset & Chengdu & Xi'an \\
    \hline 
    Time span & 09/30 - 10/10, 2018 & 09/29 - 10/15, 2018 \\
    \#Segments & 4,315 & 3,392 \\
    \#Trajectories & 140,000 & 210,000 \\
    \#Records & 18,832,411 & 18,267,440 \\
    \bottomrule
\end{tabular}

    \label{tab:dataset}
\end{table}

\subsection{Comparison Methods}
We compare the proposed method with several state-of-the-art trajectory learning methods.

\begin{itemize}[leftmargin=*]
\item \textbf{Traj2vec}~\cite{traj2vec} calculates features with sliding windows and trains the model with an auto-regressive pretext task.
\item \textbf{T2vec}~\cite{t2vec} pre-trains the model by reconstructing original trajectories from low-sampling ones using a denoising auto-encoder.
\item \textbf{TremBR}~\cite{trembr} constructs an RNN-based seq2seq model that recovers the road segments and time of the input trajectories.
\item \textbf{CTLE}~\cite{ctle} pre-trains a bi-directional Transformer with two MLM tasks of location and hour predictions. The trajectory representation is obtained by applying mean pooling on point embeddings.
\item \textbf{Toast}~\cite{toast} utilizes a context-aware node2vec model to generate segment representations and trains the model with an MLM-based task and a sequence discrimination task.
\item \textbf{TrajCL}~\cite{trajcl} introduces a dual-feature, self-attention-based encoder and trains the model in a contrastive style using the InfoNCE loss.
\item \textbf{START}~\cite{start} includes a time-aware trajectory encoder and a GAT that considers the transfer between road segments. The model is trained with both an MLM task and a contrastive task based on SimCLR loss.
\item \textbf{LightPath}~\cite{lightpath} constructs a sparse path encoder and trains it with a path reconstruction task and a cross-view \& cross-network contrastive task.
\end{itemize}

\begin{table*}[!t]
    \caption{Overall performance of methods.}
    \label{tab:overall-comparison}
    % \begin{threeparttable}
% \resizebox{1.0\linewidth}{!}{
% \begin{tabular}{c|cccccc|cccccc}
% \toprule
% Dataset & \multicolumn{6}{c|}{Chengdu} & \multicolumn{6}{c}{Xi'an} \\
% \midrule
% Downstream Task & \multicolumn{3}{c|}{Travel Time Estimation} & \multicolumn{3}{c|}{Destination Prediction} & \multicolumn{3}{c|}{Travel Time Estimation} & \multicolumn{3}{c}{Destination Prediction} \\
% \midrule
% Methods & RMSE & MAE & MAPE(\%) $\uparrow$ & ACC@1(\%) $\uparrow$ & ACC@5(\%) $\uparrow$ & Recall(\%) $\uparrow$ & RMSE & MAE & MAPE(\%) $\uparrow$ & ACC@1(\%) $\uparrow$ & ACC@5(\%) $\uparrow$ & Recall(\%) $\uparrow$ \\
% \midrule
% Traj2vec &  &  &  &  &  &  &  &  &  &  &  & \\
% T2vec &  &  &  &  &  &  &  &  &  &  &  & \\
% TremBR &  &  &  &  &  &  &  &  &  &  &  & \\
% CTLE &  &  &  &  &  &  &  &  &  &  &  & \\
% Toast&  &  &  &  &  &  &  &  &  &  &  & \\
% TrajCL&  &  &  &  &  &  &  &  &  &  &  & \\
% START &  &  &  &  &  &  &  &  &  &  &  & \\
% LightPath&  &  &  &  &  &  &  &  &  &  &  & \\
% TrajCogn(Ours)&  &  &  &  &  &  &  &  &  &  &  & \\
% \bottomrule
% \end{tabular}
% }
% \begin{tablenotes}\footnotesize
%     \item[]{\textbf{Bold} denotes the best result, and \underline{underline} denotes the second-best result.}
% \end{tablenotes}
% \end{threeparttable}

\begin{threeparttable}
\resizebox{1 \linewidth}{!}{
\begin{tabular}{c|c|ccc|ccc|ccc}
\toprule
\multicolumn{2}{c|}{Task} & \multicolumn{3}{c|}{Travel Time Estimation} & \multicolumn{3}{c|}{Destination Prediction} & \multicolumn{3}{c}{Similar Trajectory Search}\\
\midrule
Datasets & Methods & RMSE (sec) $\downarrow$ & MAE (sec) $\downarrow$ & MAPE (\%) $\downarrow$ & ACC@1 (\%) $\uparrow$ & ACC@5 (\%) $\uparrow$ & Recall (\%) $\uparrow$ & Mean Rank $\downarrow$ & ACC@1 (\%) $\uparrow$ & ACC@5 (\%) $\uparrow$ \\
\midrule
\multirow{9}{*}{Chengdu} & Traj2vec  & 130.872 $\pm$ 2.013 & 59.993 $\pm$ 2.225 & 14.870 $\pm$ 0.698 & 43.074 $\pm$ 1.255 & 73.899 $\pm$ 1.568 & 14.760 $\pm$ 0.345 & 3.371 $\pm$ 0.156 & 83.325 $\pm$ 0.754 & 89.375 $\pm$ 0.459 \\
& T2vec  & 128.508 $\pm$ 2.600 & 60.520 $\pm$ 2.575 & 15.224 $\pm$ 0.446 & 47.739 $\pm$ 0.239 & 73.509 $\pm$ 0.147 & 16.638 $\pm$ 0.108 & 3.345 $\pm$ 0.380 & 81.450 $\pm$ 0.778 & 93.700 $\pm$ 1.838 \\
& TremBR & 125.535 $\pm$ 2.849 & 57.965 $\pm$ 2.588 & 13.964 $\pm$ 0.860 & 48.987 $\pm$ 0.377 & 72.082 $\pm$ 0.289 & 17.010 $\pm$ 0.495 & 4.659 $\pm$ 1.010 & 83.980 $\pm$ 1.145 & 89.880 $\pm$ 0.303 \\
& CTLE & 132.636 $\pm$ 3.973 & 57.481 $\pm$ 1.144 & 13.153 $\pm$ 0.750 & 51.004 $\pm$ 0.683 & 79.434 $\pm$ 0.641 & 21.467 $\pm$ 0.704 & 9.429 $\pm$ 1.587 & 53.767 $\pm$ 7.414 & 69.200 $\pm$ 4.508 \\
& Toast & 128.793 $\pm$ 2.566 & 60.997 $\pm$ 3.537 & 14.883 $\pm$ 0.576 & 50.897 $\pm$ 0.495 & 79.664 $\pm$ 0.498 & 21.068 $\pm$ 0.383 & 5.944 $\pm$ 1.130 & 53.640 $\pm$ 2.244 & 71.600 $\pm$ 2.819 \\
& TrajCL & 120.211 $\pm$ 1.040 & 59.816 $\pm$ 1.841 & 14.741 $\pm$ 0.443 & 50.847 $\pm$ 0.249 & 79.693 $\pm$ 0.577 & 21.572 $\pm$ 0.324 & 1.198 $\pm$ 0.219 & 95.125 $\pm$ 5.022 & 98.875 $\pm$ 1.350 \\
& START & 122.205 $\pm$ 3.181 & 55.922 $\pm$ 2.397 & 12.717 $\pm$ 0.788 & \underline{52.775 $\pm$ 0.311} & \underline{80.423 $\pm$ 0.409} & \underline{23.316 $\pm$ 0.310} & \underline{1.089 $\pm$ 0.041} & \underline{96.933 $\pm$ 2.060} & \underline{99.900 $\pm$ 0.100}\\
% & LightPath &  &  &  & 49.154 $\pm$ & 79.587 $\pm$ 0.583 & 20.660 $\pm$ 0.272 \\
& LightPath & \underline{119.23 $\pm$ 2.367} & \underline{55.614 $\pm$ 1.518} & \underline{12.760 $\pm$ 0.854} & 49.154 $\pm$ 0.234 & 78.587 $\pm$ 0.583 & 20.660 $\pm$ 0.273 & 27.266 $\pm$ 3.544 & 74.267 $\pm$ 4.765 & 86.100 $\pm$ 3.874 \\
& \textbf{TrajCogn (ours)} & \textbf{115.079 $\pm$ 1.608} & \textbf{51.973 $\pm$ 1.922} & \textbf{11.635 $\pm$ 0.587} & \textbf{59.594 $\pm$ 0.867} & \textbf{86.740 $\pm$ 0.294} & \textbf{30.184 $\pm$ 0.875} &\textbf{ 1.068 $\pm$ 0.044} & \textbf{99.240 $\pm$ 0.152} & \textbf{99.940 $\pm$ 0.060} \\
\midrule
\multirow{9}{*}{Xi'an} & Traj2vec  & 187.010 $\pm$ 1.100 & 86.450 $\pm$ 2.884 & 13.634 $\pm$ 0.651 & 42.506 $\pm$ 0.394 & 75.761 $\pm$ 0.506 & 13.961 $\pm$ 0.376 & 2.284 $\pm$ 0.359 & 90.600 $\pm$ 0.704 & 98.017 $\pm$ 0.523\\
& T2vec  & 199.132 $\pm$ 2.447 & 86.008 $\pm$ 2.827 & 14.222 $\pm$ 0.495 & 43.596 $\pm$ 0.133 & 74.670  $\pm$ 0.343 & 13.527 $\pm$ 0.103 & 1.600 $\pm$ 0.340 & 89.467 $\pm$ 3.556 & 97.100 $\pm$ 1.637 \\
& TremBR & 185.727 $\pm$ 3.563 & 81.119 $\pm$ 2.411 & 12.770 $\pm$ 0.766 & 44.500 $\pm$ 0.349 & 75.111 $\pm$ 0.667 & 12.903 $\pm$ 0.741 & 3.478 $\pm$ 0.959 & 88.000 $\pm$ 1.355 & 93.000 $\pm$ 0.639\\
& CTLE & 182.278 $\pm$ 2.665 & \underline{79.712 $\pm$ 1.621} & 12.780 $\pm$ 0.571 & 44.837 $\pm$ 0.720 & 76.777 $\pm$ 0.610 & 14.826 $\pm$ 0.408 & 6.045 $\pm$ 1.149 & 41.200 $\pm$ 3.832 & 59.800 $\pm$ 9.835 \\
& Toast & 183.092 $\pm$ 3.827 & 84.925 $\pm$ 2.472 & 13.436 $\pm$ 0.627 & 45.078 $\pm$ 0.517 & 77.651 $\pm$ 0.123 & 15.459 $\pm$ 0.547 & 6.176 $\pm$ 1.042 & 30.600 $\pm$ 5.597 & 64.300 $\pm$ 6.505 \\
& TrajCL & \underline{179.806 $\pm$ 3.298} & 82.494 $\pm$ 2.909 & 13.231 $\pm$ 0.270 & 45.807 $\pm$ 0.474 & 79.063 $\pm$ 0.596 & \underline{16.836 $\pm$ 0.884} & \underline{1.091 $\pm$ 0.024} & 95.625 $\pm$ 1.212 & 99.200 $\pm$ 0.116
 \\
& START & 182.346 $\pm$ 3.254 & 80.763 $\pm$ 2.756 & 12.547 $\pm$ 0.501 & \underline{46.127 $\pm$ 0.267} & \underline{79.335 $\pm$ 0.489} & 16.306 $\pm$ 1.359 & 1.139 $\pm$ 0.201 & \underline{95.925 $\pm$ 3.877} & \underline{99.525 $\pm$ 0.763} \\
& LightPath & 180.032 $\pm$ 2.367 & 80.420 $\pm$ 2.189 & \underline{12.253 $\pm$ 0.686} & 44.390 $\pm$ 0.247 & 72.753 $\pm$ 0.466 & 14.416 $\pm$ 0.539 &13.877 $\pm$ 1.231 & 79.625 $\pm$ 3.236 & 91.700 $\pm$ 3.135\\
& \textbf{TrajCogn (ours)} & \textbf{166.884 $\pm$ 1.843} & \textbf{77.285 $\pm$ 2.086} & \textbf{11.357 $\pm$ 0.317} & \textbf{49.192 $\pm$ 0.238} & \textbf{81.763 $\pm$ 1.246} & \textbf{20.753 $\pm$ 0.210} & \textbf{1.083 $\pm$ 0.012} & \textbf{99.400 $\pm$ 0.254} & \textbf{99.800 $\pm$ 0.152}\\
\bottomrule
\end{tabular}
}
\begin{tablenotes}\footnotesize
    \item[]{\textbf{Bold} denotes the best result, and \underline{underline} denotes the second-best result. $\uparrow$ means higher is better, and $\downarrow$ means lower is better.}
\end{tablenotes}
\end{threeparttable}
\end{table*}

\subsection{Settings}
For each dataset, we divide the trajectories into training, validation, and testing sets in an 8:1:1 ratio, with their departure times in chronological order. Models are trained on the training set and evaluated on the testing set. The cross-reconstruction pretext task and embedding methods are pre-trained for 20 epochs, while the downstream predictors are stopped early based on the validation set. The final metrics are calculated on the testing set. We use root mean square error (RMSE), mean absolute error (MAE), and mean absolute percentage error (MAPE) for the travel time estimation task; and Top-$N$ accuracy (ACC@$N$, $N=1, 5$) and macro-F1 for the destination prediction task.

All models are implemented using PyTorch~\cite{pytorch}. We choose GPT2~\cite{gpt2} as the foundation LLM to develop our model and obtain addresses and names of POIs using Amap APIs\footnote{https://lbs.amap.com/}. The four key hyper-parameters of TrajCogn and their optimal values are $N_A=15$, $K=5$, $r=8$, and $N_\mathrm{POI}=3$. We choose parameters based on the Acc@1 and Recall of the destination prediction task on the validation set of the Chengdu dataset. We report the effectiveness of these parameters in the subsequent section. For model training, we utilize the Adam optimizer with an initial learning rate of 1e-4 for the proposed method and 0.001 for other methods. The experiments are conducted on Ubuntu 22.04 servers equipped with Intel(R) Xeon(R) W-2155 CPUs and nVidia(R) TITAN RTX GPUs. We run each set of experiments 5 times and report the mean and deviation of the metrics.

% Additionally, a visualization of the attention map between trajectory points and anchor words in the pattern semantic projector can be found in Appendix~\ref{sec:case-study}.

\begin{figure}
    \centering
    \pgfplotstableread[row sep=\\,col sep=&]{
Epoch & TrajCogn0.2 & TrajCogn0.4 & TrajCogn0.6        & TrajCogn0.8        & TrajCogn1        & START0.2 & START0.4 & START0.6           & START0.8           & START1             \\
% 0     & 0  & 0  & 0  & 0 & 0 & 0  & 0  & 0  & 0 & 0 \\
1     & 0.310369333  & 0.363399667  & 0.3983431952662720  & 0.38665009763891400 & 0.429565464356717 & 0.267045  & 0.286932  & 0.3077297034516290  & 0.32944393801276200 & 0.34694472801516700 \\
2     & 0.375710333  & 0.407078333  & 0.44189349112426000 & 0.43866500976389100 & 0.467672157497671 & 0.31108   & 0.341264  & 0.3677685950413220  & 0.3936189608021880  & 0.40790433134023600 \\
3     & 0.400568333  & 0.433830333  & 0.45633136094674600 & 0.46280845020415400 & 0.495128467224823 & 0.350142  & 0.375     & 0.40738940204180800 & 0.4282588878760260  & 0.4420300422925480  \\
4     & 0.423059     & 0.438683667  & 0.4972781065088760  & 0.48570921356293300 & 0.518750638923005 & 0.367188  & 0.39098   & 0.43485658726300400 & 0.4536007292616230  & 0.46988478926644300 \\
5     & 0.442235     & 0.468039667  & 0.49775147928994100 & 0.49795845908041900 & 0.521356054017949 & 0.365767  & 0.414418  & 0.4404472532814780  & 0.47748404740200500 & 0.4875309902289630  \\
6     & 0.448863667  & 0.482007667  & 0.5188165680473370  & 0.5084324516243570  & 0.539014011004402 & 0.356534  & 0.417614  & 0.4581915410792420  & 0.49626253418413900 & 0.5041563365903460  \\
7     & 0.451941333  & 0.499763667  & 0.5183431952662720  & 0.5123380081661640  & 0.547583640811076 & 0.364347  & 0.427202  & 0.4649975692756440  & 0.5031905195989060  & 0.5149482280880850  \\
8     & 0.456912667  & 0.496922333  & 0.5263905325443790  & 0.5219243742233270  & 0.55260314281315  & 0.379972  & 0.440696  & 0.4829849295089940  & 0.5124886052871470  & 0.5293860288756020  \\
9     & 0.472064333  & 0.497514333  & 0.5453254437869820  & 0.5323983667672640  & 0.552866041577506 & 0.356534  & 0.422585  & 0.46937287311618900 & 0.5110300820419330  & 0.5270526469301440  \\
10    & 0.467329667  & 0.506747333  & 0.5363313609467460  & 0.5386117521746850  & 0.56207412636856  & 0.367188  & 0.427912  & 0.48565872630043800 & 0.5157702825888790  & 0.5325944290506050  \\
11    & 0.476325667  & 0.509706333  & 0.5342011834319530  & 0.5327534173619740  & 0.562123449444192 & 0.365767  & 0.424716  & 0.4832280019445800  & 0.5205104831358250  & 0.526323465072189   \\
12    & 0.478456333  & 0.514441333  & 0.5398816568047340  & 0.5522812000710100  & 0.559474675388327 & 0.355114  & 0.427557  & 0.47763733592610600 & 0.5192342752962630  & 0.5194691556074090  \\
13    & 0.470407333  & 0.516098333  & 0.5427218934911240  & 0.5442925616900410  & 0.566624254073161 & 0.362926  & 0.428622  & 0.4747204666990760  & 0.5217866909753870  & 0.5311360653346950  \\
14    & 0.490767333  & 0.523556     & 0.5427218934911240  & 0.5469554411503640  & 0.572353781636155 & 0.361506  & 0.428267  & 0.48079727758872100 & 0.515587967183227   & 0.5257401195858250  \\
15    & 0.491950667  & 0.518584333  & 0.5493491124260360  & 0.5453577134741700  & 0.57112505870213  & 0.385653  & 0.426847  & 0.47593582887700500 & 0.5026435733819510  & 0.5136357007437660  \\
16    & 0.480587333  & 0.517992333  & 0.5512426035502960  & 0.5547665542339780  & 0.582392785640301 & 0.348722  & 0.42152   & 0.47520661157024800 & 0.5026435733819510  & 0.5216567011812750  \\
17    & 0.484848667  & 0.524976333  & 0.5571597633136100  & 0.5554766554233980  & 0.578891980911332 & 0.357923  & 0.419389  & 0.48128342245989300 & 0.5110300820419330  & 0.5318652471926500  \\
18    & 0.478693333  & 0.522964     & 0.554792899408284   & 0.5576069589916560  & 0.582065416455013 & 0.367188  & 0.420099  & 0.4637822070977150  & 0.5081130355515040  & 0.5289485197608280  \\
19    & 0.480823667  & 0.529119333  & 0.549585798816568   & 0.5570743830995920  & 0.586232887783982 & 0.362216  & 0.430398  & 0.47520661157024800 & 0.5044667274384690  & 0.5273443196733270  \\
20    & 0.486032333  & 0.517992333  & 0.5486390532544380  & 0.5599147878572700  & 0.577903909240756 & 0.339489  & 0.430398  & 0.46985901798736000 & 0.5044667274384690  & 0.5215108648096840  \\
}\Scalability

\newcommand{\accMin}{26}
\newcommand{\accMax}{61}
\newcommand{\accTick}{10}

\tikzset{every plot/.style={line width=0.8pt}}

\ref{fig:scalability-legend}

\begin{subfigure}[b]{\linewidth}
    \begin{tikzpicture}
    \begin{axis}[
        width=1\linewidth, height=0.6\linewidth,
        ylabel={ACC@1(\%)}, 
        ylabel style={at={(axis description cs:0.,0.5)},anchor=north},
        ymin=\accMin, ymax=\accMax,
        ytick distance={\accTick},
        ymajorgrids=true,
        grid style=dashed,
        % xtick=data,
        % xticklabels={0,5,10,15,20},
        % symbolic x coords={0, 10, 20},
        % legend entries={TrajCogn 1, TrajCogn 0.6, START 0.6， TrajCogn 0.2, START 0.2 },
        legend to name=fig:scalability-legend,
        legend columns=3,
    ]
    \addplot[color=prop1,mark=square*, opacity=0.8] table[x=Epoch,y expr=\thisrow{TrajCogn1} * 100]{\Scalability};
    \addlegendentry{TrajCogn,1}

    \addplot[color=prop0.6,mark=square*, opacity=0.8] table[x=Epoch,y expr=\thisrow{TrajCogn0.6} * 100]{\Scalability};
    \addlegendentry{TrajCogn,0.6}

    \addplot[color=prop0.2,mark=square*, opacity=0.8] table[x=Epoch,y expr=\thisrow{TrajCogn0.2} * 100]{\Scalability};
    \addlegendentry{TrajCogn,0.2}
    
    \addplot[color=prop1,mark=triangle*, opacity=0.8] table[x=Epoch,y expr=\thisrow{START1}*100]{\Scalability};
    \addlegendentry{START,1}
    
    \addplot[color=prop0.6,mark=triangle*, opacity=0.8] table[x=Epoch,y expr=\thisrow{START0.6}*100]{\Scalability};
    \addlegendentry{START,0.6}
    
    \addplot[color=prop0.2,mark=triangle*, opacity=0.8] table[x=Epoch,y expr=\thisrow{START0.2}*100]{\Scalability};
    \addlegendentry{START,0.2}
    \addlegendimage{/pgfplots/refstyle=recallstyle}
    \end{axis}
    % \begin{axis}[
    %     width=1.15\linewidth, height=1.0\linewidth,
    %     % axis y line*=right,
    %     axis x line=none, 
    %     % ylabel={Recall(\%)},
    %     % ylabel near ticks,
    %     % ymin=\recMin, ymax=\recMax,
    %     % ytick distance={\recTick},
    %     ymajorgrids=true,
    %     grid style=dashed,
    %     xtick=data, 
    %     % symbolic x coords={0, 10, 20}
    % ]
    % \addplot[color=recall,mark=triangle*] table[x=Epoch,y expr=\thisrow{START1}*100]{\Scalability};
    % \label{recallstyle}
    % \end{axis}
    \end{tikzpicture}
    % \caption{Scalability}
    % \label{fig:scalability}
\end{subfigure}
    \caption{Scalability of fine-tuning on Chengdu.}
    \label{fig:scalability}
\end{figure}

\begin{table}
    \caption{Efficiency of methods on Chengdu.}
    \resizebox{1 \linewidth}{!}{\begin{threeparttable}
% \resizebox{1 \linewidth}{!}{
\begin{tabular}{c|cccc}
    \toprule
     \multirow{2}{*}{Methods} & Learnable & Pre-Train Speed & Fine-Tune Speed & Embed  \\
     & Param (MB) & (min/epoch) & (min/epoch) & Time (sec) \\
    \midrule
    CTLE & 3.756 & 4.533 & 3.516 & 14.581 \\
    Toast & 4.007 & 4.400 & 3.517 & 14.539 \\
    TrajCL & 5.634 & 7.699 & 4.543 & 10.253 \\
    START & 15.928 & 15.927 & 7.573 & 28.704 \\
    \textbf{TrajCogn} & 27.922 & 24.931 & 19.644 & 110.516 \\
    \bottomrule
\end{tabular}
% }
\end{threeparttable}}

    \label{tab:efficiency}
\end{table}

\subsection{Performance Comparison}
\subsubsection{Overall Performance}
Table~\ref{tab:overall-comparison} presents a comprehensive comparison of the performance of all task-adaptable trajectory learning methods across two tasks and two datasets. Our proposed method consistently outperforms the others and performs well across tasks, providing evidence that it is an advanced task-adaptable trajectory learning method.

Traj2vec, T2vec, and TremBR all adopt RNN-based auto-encoding or auto-regressive frameworks. T2vec and TremBR do not consider crucial spatio-temporal features and cannot capture travel purposes, resulting in subpar performance on downstream tasks. CTLE and Toast use bi-directional Transformers and incorporate MLM tasks~\cite{bert}. Their performance suffers due to the absence of essential continuous features and the inability to extract travel purposes. Notably, their performance on the STS task is suboptimal as they do not learn a trajectory-level representation directly.

TrajCL, START, and LightPath employ contrastive learning pretext tasks, which contribute to their performance in the STS task. However, these methods fail to consider the functionalities of POIs, and START and LightPath also face challenges in extracting movement patterns due to insufficient consideration of continuous features. Consequently, they do not yield satisfactory results in TTE and DP tasks.

Our proposed method leverages the strong capabilities of LLMs for task-adaptable trajectory learning and can be adapted to various downstream tasks, regardless of the size of trajectory datasets. It effectively extracts movement patterns and provides explainability using the power of the LLM. The proposed model preserves the inherent functionalities of POIs around the OD points and incorporates travel purposes by employing a trajectory prompt that includes POIs. These advantages contribute to the superior performance of our model in multiple downstream tasks.

\subsubsection{Scalability}
To compare the scalability of the proposed model against START, one of the state-of-the-art models, we refine our model using varying proportions of the training data: 100\%, 60\%, and 20\% for the destination prediction task on the Chengdu dataset. We use the START model as a reference point with an identical learning rate of $5 \times 10^{-4}$ for comparison. The results are presented in Figure~\ref{fig:scalability}. It can be seen that our model demonstrates faster progress and achieves superior performance with less data compared to START. This shows that our model can be adapted to downstream tasks with lightweight finetuning.

\begin{table*}
    \caption{Performance of variants of TrajCogn.}
    \label{tab:ablation}
    \begin{threeparttable}
\resizebox{1.0\linewidth}{!}{
\begin{tabular}{c|ccc|ccc|ccc}
\toprule
Task & \multicolumn{3}{c|}{Travel Time Estimation} & \multicolumn{3}{c|}{Destination Prediction} 
 & \multicolumn{3}{c}{Similar Trajectory Search} \\
\midrule
Methods & RMSE (sec) $\downarrow$ & MAE (sec) $\downarrow$ & MAPE (\%) $\downarrow$ & ACC@1 (\%) $\uparrow$ & ACC@5 (\%) $\uparrow$ & Recall (\%) $\uparrow$ & Mean Rank $\downarrow$ & ACC@1 (\%) $\uparrow$ & ACC@5 (\%) $\uparrow$ \\
\midrule
w/o PT & 120.737 $\pm$ 0.634 & 54.951 $\pm$ 2.632 & 12.087 $\pm$ 0.980 & 57.455 $\pm$ 0.723 & 85.331 $\pm$ 0.161 & 28.390 $\pm$ 1.512  & 3.914 $\pm$ 0.033 & 88.000 $\pm$ 0.566 & 94.600 $\pm$ 0.707\\
w/o POI & 116.132 $\pm$ 2.131 & 52.941 $\pm$ 4.453 & 12.080 $\pm$ 0.924 & 58.711 $\pm$ 0.215 & 86.128 $\pm$ 0.118 & 29.372 $\pm$ 0.666 & 1.092 $\pm$ 0.065 & 98.200 $\pm$ 2.115 & 99.325 $\pm$ 0.754 \\
w/o Conv & 117.038 $\pm$ 2.237 & 53.402 $\pm$ 3.175 & \underline{11.836 $\pm$ 1.175} & \underline{59.078 $\pm$ 1.054} & 86.200 $\pm$ 0.673 & 29.521 $\pm$ 1.477 &1.137 $\pm$ 0.050 & 96.733 $\pm$ 1.823 & 98.700 $\pm$ 0.781
\\
w/o PSP & 115.454 $\pm$ 5.551 & 53.003 $\pm$ 2.363 & 12.265 $\pm$ 0.856 & 58.797 $\pm$ 0.698 & 86.166 $\pm$ 0.460 & 29.503 $\pm$ 0.779 & 1.256 $\pm$ 0.256 & 96.667 $\pm$ 2.214 & 98.367 $\pm$ 1.037\\
w/o $\mathcal{M}$ & \underline{115.233 $\pm$ 0.509} & \underline{52.790 $\pm$ 3.297} & 11.891 $\pm$ 0.794 & 58.930 $\pm$ 0.220 & \underline{86.668 $\pm$ 0.324} & \underline{29.626 $\pm$ 0.287} & \underline{1.069 $\pm$ 0.022} & \underline{98.525 $\pm$ 0.551} & \underline{99.350 $\pm$ 0.100}\\
TrajCogn (full) & \textbf{115.079 $\pm$ 1.608} & \textbf{51.973 $\pm$ 1.922} & \textbf{11.635 $\pm$ 0.587} & \textbf{59.594 $\pm$ 0.867} & \textbf{86.740 $\pm$ 0.294} & \textbf{30.184 $\pm$ 0.875} & \textbf{1.068 $\pm$ 0.044} & \textbf{99.240 $\pm$ 0.152} & \textbf{99.940 $\pm$ 0.060} \\
\bottomrule
\end{tabular}
}
\begin{tablenotes}\footnotesize
    \item[]{\textbf{Bold} denotes the best result, and \underline{underline} denotes the second-best result. $\uparrow$ means higher is better, and $\downarrow$ means lower is better.}
\end{tablenotes}
\end{threeparttable}
\end{table*}

\subsubsection{Efficiency}
We investigate the efficiency of TrajCogn in comparison to other methods. We conduct the comparison on the Chengdu dataset and set the batch size of both pre-training and fine-tuning as 16. The result is detailed in Table~\ref{tab:efficiency}. We consider the learnable parameter scale, pre-training speed, fine-tuning speed, and embedding time on the Chengdu dataset. While incorporating PLMs increases the model scale and reduces training speed, we have implemented efficient adaptation strategies such as LoRA to ensure that the additional learnable parameters and training speed remain reasonable.

\subsection{Model Analysis}
\subsubsection{Effectiveness of Components}
To assess the effectiveness of the components implemented in TrajCogn, we compared the performance of the complete model with the following variants:
\begin{enumerate}[leftmargin=*]
\item \textit{w/o PT} removes the cross-reconstruction pretext task and trained the model directly on downstream tasks.
\item \textit{w/o POI} removes the $\langle$POI Part$\rangle$ from the trajectory prompt.
\item \textit{w/o Conv} removes the convolution operator in the trajectory semantic embedder and used a one-layer fully connected layer for continuous feature embedding.
\item \textit{w/o PSP} removes the pattern semantic projector and used $\boldsymbol{e}_i$ from Equation~\ref{equ:fea-embedding} as the trajectory point embedding $\boldsymbol{z}_i$.
\item \textit{w/o $\mathcal{M}$} removes the movement pattern vocabulary $M$ and only used the virtual anchor words in the pattern semantic projector.
\end{enumerate}

We measured the performance of these variants on the Chengdu dataset, and the results are presented in Table~\ref{tab:ablation}. Based on the results, we made the following observations:
\begin{enumerate}[leftmargin=*]
\item \textit{w/o PT} shows performance degradation, proving the contribution of the cross-reconstruction pretext task to TrajCogn.
\item The worse performance witnessed by \textit{w/o POI} demonstrates the effectiveness of integrating POI information.
\item \textit{w/o Conv}, \textit{w/o PSP}, and \textit{w/o $\mathcal{M}$} all have worse performance compared to \textit{full}, showing that the removed components all contribute to TrajCogn's performance.
\end{enumerate}

\begin{table}
    \centering
    \caption{Performance of anchor word selection strategies.}
    \begin{threeparttable}
\resizebox{1 \linewidth}{!}{
\begin{tabular}{c|ccc}
\toprule
Variants & ACC@1 (\%) & ACC@5 (\%) & Recall (\%) \\
\midrule
w/o $\mathcal{M}$ & 58.930 $\pm$ 0.220 & 86.668 $\pm$ 0.324 & 29.626 $\pm$ 0.287 \\
Decrease & 59.191 $\pm$ 0.291 & 86.791 $\pm$ 0.424 & 29.776 $\pm$ 0.439 \\
Replace & 58.107 $\pm$ 0.329 & 85.948 $\pm$ 0.237 & 28.798 $\pm$ 0.697 \\
TrajCogn &\textbf{59.594 $\pm$ 0.867} & \textbf{86.740 $\pm$ 0.294} & \textbf{30.184 $\pm$ 0.875} \\
\bottomrule
\end{tabular}
}
% \begin{tablenotes}\footnotesize
    % \item[]{\textbf{Bold} denotes the best result, and \underline{underline} denotes the second-best result.}
% \end{tablenotes}
\end{threeparttable}
    \label{tab:word-selection}
\end{table}

\subsubsection{Impact of Anchor Word Selection}
\label{sec:anchor-word-selection}
To investigate the impact of anchor words used in the movement pattern semantic projection, we compare the performance of the current selection strategy with the following variants:
\begin{itemize}[leftmargin=*]
    \item \textit{w/o $\mathcal{M}$} excludes the movement pattern vocabulary $M$, retaining only virtual anchor words within the pattern semantic projector.
    \item \textit{Decrease} reduces the predefined movement pattern descriptive words to half their original number.
    \item \textit{Replace} substitutes the anchor words with an equal number of adjectives unrelated to movement patterns, such as "good", "new", and "little".
\end{itemize}

We measure the performance of these variants on the destination prediction task using the Chengdu dataset, and the results are presented in Table~\ref{tab:word-selection}. It can be observed that reducing vocabulary and replacing it with irrelevant words both lead to worse results, proving the rationality of our selection strategy.

\subsubsection{Additional Model Analysis}
We present additional model analysis in the Appendix to provide further insights into the effectiveness of TrajCogn. Specifically, we analyze the effectiveness of hyper-parameters in Appendix~\ref{sec:hyper-params}, investigate the impact of additional features in Appendix~\ref{sec:additional-features}, and present a visualization of the attention map between trajectory points and anchor words in Appendix~\ref{sec:case-study}.

\section{Conclusion}
We propose TrajCogn, a novel trajectory learning model that leverages LLMs to model trajectories and accurately perform various trajectory-related tasks. TrajCogn introduces a trajectory prompt that integrates two key aspects of information: movement patterns and travel purposes. This prompt also enables the model to adapt to different tasks. Additionally, TrajCogn includes a trajectory semantic embedder, allowing LLMs to process the spatio-temporal features of trajectories. This facilitates the effective and explainable extraction of movement patterns and travel purposes. Experimental results on two real-world datasets in various settings demonstrate the superior performance of TrajCogn.

\begin{acks}
This work was supported by the National Natural Science Foundation of China (No. 62372031).
\end{acks}

\newpage

\appendix

\section{STS Ground Truth Construction}
\label{sec:sts-gt-construction}
We randomly select 1,000 trajectories from the test dataset. For each trajectory $\mathcal{T}$, we collect the odd-numbered points to form the query $\mathcal{T}^q$ and the even-numbered points to create the target $\mathcal{T}^t$. For each query, we discard the top 10 trajectories that are closest to the query, and then randomly choose 5,000 additional trajectories from the rest of the test dataset to serve as the database. In calculating the distances between the query and other trajectories, we follow \cite{DBLP:conf/kdd/Fang0ZHCGJ22}, downsampling them to a uniform length and then computing the mean square error.

\begin{table*}
    \centering
    \caption{Performance of variants of TrajCogn and START on Chengdu.}
    \begin{threeparttable}
% \resizebox{1 \linewidth}{!}{
\begin{tabular}{c|ccc|ccc}
\toprule
{Downstream Task} & \multicolumn{3}{c|}{Destination Prediction} & \multicolumn{3}{c}{Similar Trajectory Search} \\
\midrule
Methods & ACC@1 (\%) $\uparrow$ & ACC@5 (\%) $\uparrow$ & Recall (\%) $\uparrow$ & Mean Rank $\downarrow$ & ACC@1 (\%) $\uparrow$ & ACC@5(\%) $\uparrow$ \\
\midrule
START & 52.775 $\pm$ 0.311 & 80.423 $\pm$ 0.409 & 23.316 $\pm$ 0.310 & 1.089 $\pm$ 0.041 & 96.933 $\pm$ 2.060 & \underline{99.900 $\pm$ 0.100} \\
START w/ AF & 53.287 $\pm$ 0.172 & 81.897 $\pm$ 0.191 & 23.897 $\pm$ 0.321 & 1.073 $\pm$ 0.006 & 96.200 $\pm$ 0.707 & 99.850 $\pm$ 0.071\\
\makecell[c]{TrajCogn w/o AF} & \underline{56.565 $\pm$ 0.360} & \underline{85.023 $\pm$ 0.176} & \underline{27.833 $\pm$ 0.302} & \underline{1.072 $\pm$ 0.035} & \underline{98.600 $\pm$ 1.097} & 99.650 $\pm$ 0.336 \\
TrajCogn &\textbf{59.594 $\pm$ 0.867} & \textbf{86.740 $\pm$ 0.294} & \textbf{30.184 $\pm$ 0.875}& \textbf{1.068 $\pm$ 0.044} & \textbf{99.240 $\pm$ 0.152} & \textbf{99.940 $\pm$ 0.060} \\
\bottomrule
\end{tabular}
% }
\begin{tablenotes}\footnotesize
    \item[]{\textbf{Bold} denotes the best result, and \underline{underline} denotes the second-best result. $\uparrow$ means higher is better, and $\downarrow$ means lower is better.}
\end{tablenotes}
\end{threeparttable}
    \label{tab:additional-features}
\end{table*}

\begin{figure}
    \centering
    \pgfplotstableread[row sep=\\,col sep=&]{
v & acc@1 & acc@5 & acc@10 & macro-f1 & macro-rec & mean-rank \\
0 & 0.583389 & 0.858108 & 0.903723 & 0.2589 & 0.2831 & 31.1806 \\
15 & 0.593895 & 0.866373 & 0.911815 & 0.275055 & 0.296801 & 25.852457 \\
30 & 0.594037 & 0.866231 & 0.914087 & 0.268834 & 0.295169 & 27.179494 \\
60 & 0.594103 & 0.86645 & 0.914143 & 0.268903 & 0.295301 & 27.180123 \\
}\AnoNum

\pgfplotstableread[row sep=\\,col sep=&]{
v & acc@1 & acc@5 & acc@10 & macro-f1 & macro-rec & mean-rank \\
4 & 0.588499 & 0.859557 & 0.904431 & 0.260809 & 0.289098 & 27.208321 \\
8 & 0.594037 & 0.864385 & 0.909685 & 0.273926 & 0.298605 & 27.488214 \\ 
16 & 0.592191 & 0.862681 & 0.906135 & 0.272716 & 0.298752 & 30.933542 \\
64 & 0.583103 & 0.857995 & 0.903437 & 0.2627 & 0.2878 & 31.180631 \\
}\LoRARank

\pgfplotstableread[row sep=\\,col sep=&]{
v & acc@1 & acc@5 & acc@10 & macro-f1 & macro-rec & mean-rank \\
1 & 0.582931 & 0.857995 & 0.905425 & 0.25353 & 0.281574 & 31.15436 \\
3 & 0.588753 & 0.859273 & 0.906561 & 0.265521 & 0.293681 & 25.818091 \\
5 & 0.595427 & 0.863959 & 0.908549 & 0.2747 & 0.302799 & 29.831866 \\
7 & 0.590173 & 0.861261 & 0.907839 & 0.273999 & 0.300021 & 28.357285 \\
}\KernelSize

\pgfplotstableread[row sep=\\,col sep=&]{
v & acc@1 & acc@5 & acc@10 & macro-f1 & macro-rec & mean-rank \\
1 & 0.588215 & 0.856149 & 0.903721 & 0.255149 & 0.281334 & 30.428997 \\
3 & 0.592617 & 0.861403 & 0.904431 & 0.267352 & 0.297213 & 31.350043 \\
5 & 0.584097 & 0.861119 & 0.9023 & 0.256359 & 0.286871 & 33.399176 \\
10 & 0.577991 & 0.843368 & 0.89449 & 0.234148 & 0.264851 & 23.291679 \\
}\POINumber

\pgfplotstableread[row sep=\\,col sep=&]{
v & acc@1 & acc@5 & acc@10 & macro-f1 & macro-rec & mean-rank \\
64 & 0.590771 & 0.868391 & 0.910401 & 0.274262 & 0.30181 & 27.699375 \\
128 & 0.594735 & 0.868817 & 0.910685 & 0.274031 & 0.301153 & 27.961517 \\
256 & 0.592475 & 0.867965 & 0.909123 & 0.274204 & 0.301378 & 28.095995 \\
512 & 0.590925 & 0.864149 & 0.907999 & 0.272287 & 0.300409 & 26.788128 \\
768 & 0.597262 & 0.870917 & 0.911815 & 0.275783 & 0.304574 & 26.380574 \\
}\OutputSize

\newcommand{\accMin}{56}
\newcommand{\accMax}{63}
\newcommand{\accTick}{2}
\newcommand{\recMin}{25}
\newcommand{\recMax}{32}
\newcommand{\recTick}{2}

\tikzset{every plot/.style={line width=1.5pt}}

\ref{fig:shared-legend}

\begin{subfigure}[b]{0.48\linewidth}
    \begin{tikzpicture}
    \begin{axis}[
        width=1.15\linewidth, height=1.0\linewidth,
        ylabel={ACC@1(\%)}, 
        ylabel style={at={(axis description cs:0.1,0.5)},anchor=north},
        ymin=\accMin, ymax=\accMax,
        ytick distance={\accTick},
        ymajorgrids=true,
        grid style=dashed,
        xtick=data,
        symbolic x coords={0, 15, 30, 60},
        legend entries={ACC@1, Recall},
        legend to name=fig:shared-legend,
        legend columns=-1,
    ]
    \addplot[color=acc,mark=square*] table[x=v,y expr=\thisrow{acc@1} * 100]{\AnoNum};
    \addlegendentry{ACC@1}
    \addlegendimage{/pgfplots/refstyle=recallstyle}
    \addlegendentry{Recall}
    \end{axis}
    \begin{axis}[
        width=1.15\linewidth, height=1.0\linewidth,
        axis y line*=right,
        axis x line=none, 
        ylabel near ticks,
        ymin=\recMin, ymax=\recMax,
        ytick distance={\recTick},
        ymajorgrids=true,
        grid style=dashed,
        xtick=data, 
        symbolic x coords={0, 15, 30, 60}
    ]
    \addplot[color=recall,mark=triangle*] table[x=v,y expr=\thisrow{macro-rec}*100]{\AnoNum};
    \label{recallstyle}
    \end{axis}
    \end{tikzpicture}
    \caption{Number of virtual words $N_A$}
    \label{subfig:NA}
\end{subfigure}
\hfill
\begin{subfigure}[b]{0.48\linewidth}
    \begin{tikzpicture}
    \begin{axis}[
        width=1.15\linewidth, height=1.0\linewidth,
        ymin=\accMin, ymax=\accMax,
        ytick distance={\accTick},
        ymajorgrids=true,
        grid style=dashed,
        xtick=data,
        symbolic x coords={4, 8, 16, 64}
    ]
    \addplot[color=acc,mark=square*] table[x=v,y expr=\thisrow{acc@1} * 100]{\LoRARank};
    \end{axis}
    \begin{axis}[
        width=1.15\linewidth, height=1.0\linewidth,
        axis y line*=right,
        axis x line=none, 
        ylabel={Recall(\%)},
        ylabel near ticks,
        ymin=\recMin, ymax=\recMax,
        ytick distance={\recTick},
        ymajorgrids=true,
        grid style=dashed,
        xtick=data, 
        symbolic x coords={4, 8, 16, 64}
    ]
    \addplot[color=recall,mark=triangle*] table[x=v,y expr=\thisrow{macro-rec}*100]{\LoRARank};
    \end{axis}
    \end{tikzpicture}
    \caption{LoRA rank $r$}
    \label{subfig:lora-rank}
\end{subfigure}
\hfill
\begin{subfigure}[b]{0.48\linewidth}
    \begin{tikzpicture}
    \begin{axis}[
        width=1.15\linewidth, height=1.0\linewidth,
        ylabel={ACC@1(\%)}, 
        ylabel style={at={(axis description cs:0.1,0.5)},anchor=north},
        ymin=\accMin, ymax=\accMax,
        ytick distance={\accTick},
        ymajorgrids=true,
        grid style=dashed,
        xtick=data,
        symbolic x coords={1, 3, 5, 7}
    ]
    \addplot[color=acc,mark=square*] table[x=v,y expr=\thisrow{acc@1} * 100]{\KernelSize};
    \end{axis}
    \begin{axis}[
        width=1.15\linewidth, height=1.0\linewidth,
        axis y line*=right,
        axis x line=none, 
        ylabel near ticks,
        ymin=\recMin, ymax=\recMax,
        ytick distance={\recTick},
        ymajorgrids=true,
        grid style=dashed,
        xtick=data, 
        symbolic x coords={1, 3, 5, 7}
    ]
    \addplot[color=recall,mark=triangle*] table[x=v,y expr=\thisrow{macro-rec}*100]{\KernelSize};
    \end{axis}
    \end{tikzpicture}
    \caption{Kernel Size $K$}
    \label{subfig:kernel-size}
\end{subfigure}
\hfill
\begin{subfigure}[b]{0.48\linewidth}
    \begin{tikzpicture}
    \begin{axis}[
        width=1.15\linewidth, height=1.0\linewidth,
        ymin=\accMin, ymax=\accMax,
        ytick distance={\accTick},
        ymajorgrids=true,
        grid style=dashed,
        xtick=data,
        symbolic x coords={1, 3, 5, 10}
    ]
    \addplot[color=acc,mark=square*] table[x=v,y expr=\thisrow{acc@1} * 100]{\POINumber};
    \end{axis}
    \begin{axis}[
        width=1.15\linewidth, height=1.0\linewidth,
        axis y line*=right,
        axis x line=none, 
        ylabel={Recall(\%)},
        ylabel near ticks,
        ymin=\recMin, ymax=\recMax,
        ytick distance={\recTick},
        ymajorgrids=true,
        grid style=dashed,
        xtick=data, 
        symbolic x coords={1, 3, 5, 10}
    ]
    \addplot[color=recall,mark=triangle*] table[x=v,y expr=\thisrow{macro-rec}*100]{\POINumber};
    \end{axis}
    \end{tikzpicture}
    \caption{Number of POIs $N_\text{POI}$}
    \label{subfig:poi-num}
\end{subfigure}

% \begin{subfigure}[b]{0.48\linewidth}
%     \begin{tikzpicture}
%     \begin{axis}[
%         width=1.15\linewidth, height=1.0\linewidth,
%         ylabel={ACC@1(\%)}, 
%         ylabel style={at={(axis description cs:0.1,0.5)},anchor=north},
%         ymin=\accMin, ymax=\accMax,
%         ytick distance={\accTick},
%         ymajorgrids=true,
%         grid style=dashed,
%         xtick=data,
%         symbolic x coords={64, 128, 256, 512, 768}
%     ]
%     \addplot[color=acc,mark=square*] table[x=v,y expr=\thisrow{acc@1} * 100]{\OutputSize};
%     \end{axis}
%     \begin{axis}[
%         width=1.15\linewidth, height=1.0\linewidth,
%         axis y line*=right,
%         axis x line=none, 
%         ylabel={Recall(\%)},
%         ylabel near ticks,
%         ymin=\recMin, ymax=\recMax,
%         ytick distance={\recTick},
%         ymajorgrids=true,
%         grid style=dashed,
%         xtick=data, 
%         symbolic x coords={64, 128, 256, 512, 768}
%     ]
%     \addplot[color=recall,mark=triangle*] table[x=v,y expr=\thisrow{macro-rec}*100]{\OutputSize};
%     \end{axis}
%     \end{tikzpicture}
%     \caption{Output dimension $d_O$}
%     \label{subfig:output-size}
% \end{subfigure}
    \caption{Effectiveness of hyper-parameters.}
    \label{fig:hyper-parameters}
\end{figure}

\section{Effectiveness of Hyper-parameters}
\label{sec:hyper-params}
We analyze the effectiveness of the key hyper-parameters $N_A$, $r$, $k$, and $N_\text{POI}$ on the performance of TrajCogn. We use the Acc@1 and Recall metrics of the destination prediction task. The results obtained on the Chengdu dataset are presented in Figure~\ref{fig:hyper-parameters}. We make the following observations:
\begin{enumerate}[leftmargin=*]
    \item As illustrated in Figure~\ref{subfig:NA}, increasing the number of virtual anchor words generally improves performance. However, beyond $N_A = 15$, the improvements in both accuracy and recall are negligible, while computation and memory requirements increase. Therefore, we set $N_A = 15$ to balance performance and efficiency.
    \item We set the rank in LoRA to $r=8$. As illustrated in Figure~\ref{subfig:lora-rank}, a smaller $r$ decreases model complexity, making it challenging to fit the LLM on trajectory data. In contrast, a larger rank increases model capacity, leading to overfitting.
    \item The convolution kernel with size 5 leads to optimal performance, so we set the kernel size to $k=5$. A smaller receptive field is inadequate for accurately identifying the movement pattern of the current trajectory point, while a larger receptive field results in over-smoothing of features.
    \item The number of POIs $N_\text{POI}$ has an optimal value of 3, as seen in Figure~\ref{subfig:poi-num}. A smaller number of POIs can indicate a wrong origin or destination, while more POIs may introduce more noise.
\end{enumerate}

\section{Impact of Additional Features}
\label{sec:additional-features}
To examine the impact of incorporating extra attributes such as velocity, acceleration, and direction, we omitted these features from the TrajCogn model and integrated them into START, denoted as \textit{w/o AF} and \textit{w/ AF}, respectively. We then compared the performance of both models on the DP and STS tasks using the Chengdu dataset. To prevent data leakage, we did not use additional features in the original TrajCogn for the TTE task. The results are presented in Table~\ref{tab:additional-features}. We observe that \textit{TrajCogn w/o AF} yields subpar results, proving the benefit of additional features. Additionally, \textit{START w/ AF} shows improved performance, but not exceeding TrajCogn.

\begin{figure*}
    \centering
    \includegraphics[width=\linewidth]{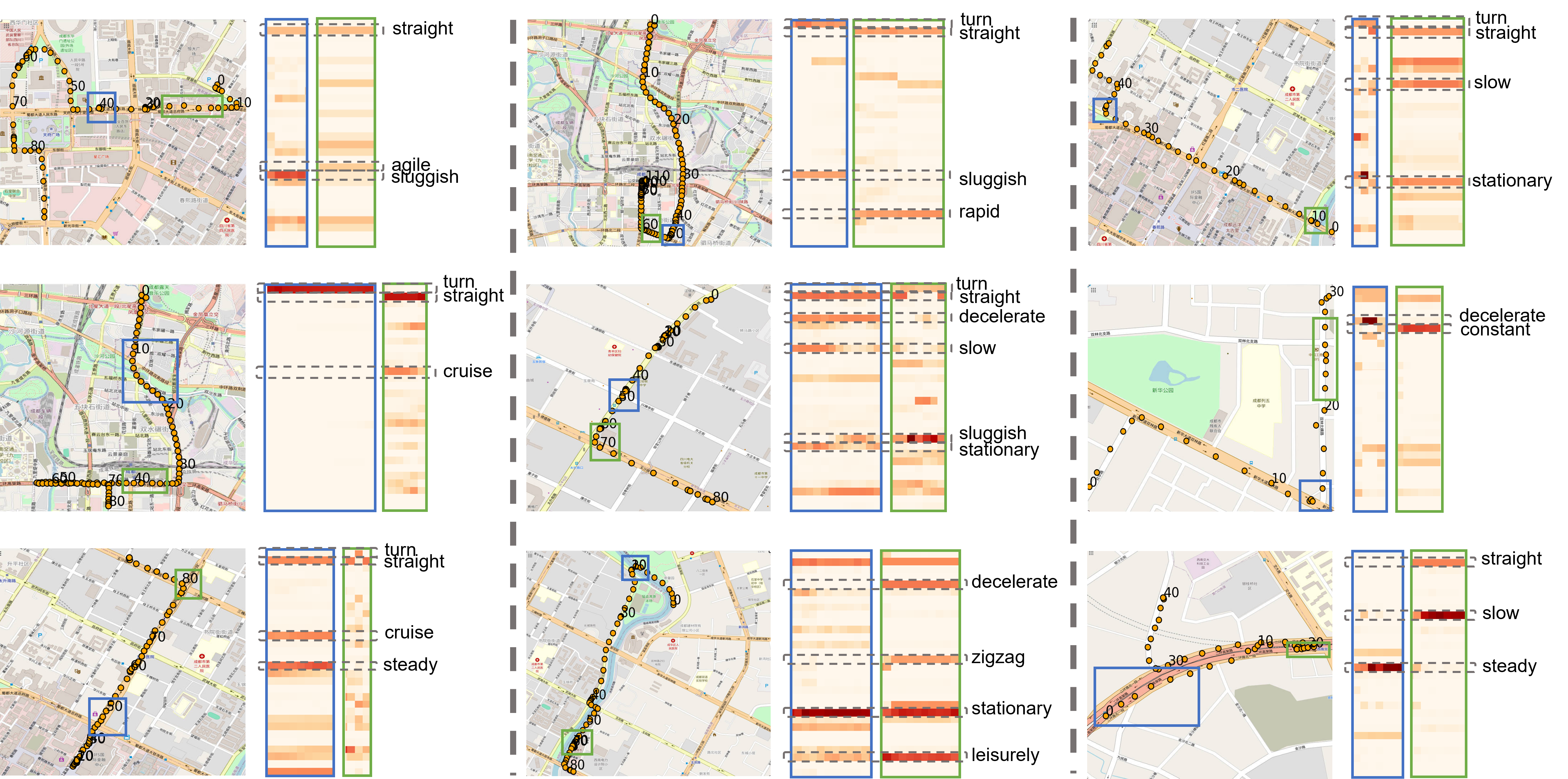}
    \caption{Visualization of attention maps in the movement pattern semantic projection.}
    \label{fig:attn-map}
\end{figure*}

\section{Attention Map Visualization}
\label{sec:case-study}
To demonstrate how our model effectively extracts movement patterns with considerable interpretability, we present an intuitive visualization of the attention scores in the pattern semantic projector, as shown in Figure~\ref{fig:attn-map}. In each example, the original trajectory is on the left side, with certain points marked by sequence indices. Two subtrajectories are highlighted by blue and green boxes. The attention maps for these subtrajectories are displayed on the right side. In this case, we set $N_A=0$ and evaluated the model's performance after training on the cross-reconstruction pretext task for 20 epochs.

We discovered that specific movement patterns displayed by trajectory points are associated with particular anchor words. Key terms such as "turn", "slow", and "steady" within these anchor words uncover the underlying semantics of the movement patterns. Upon observing Figure~\ref{fig:attn-map}, when the object makes a turn, the attention scores for "turn" increase. A high association with words like "slow", "sluggish", and "stationary" suggests the object is moving slowly. Meanwhile, a trajectory that progresses steadily is strongly correlated with terms such as "steady", "cruise", and "straight".

However, the words linked to these patterns do not always precisely convey the true semantics of the movements. Accurate labeled data is required for more precise alignment effects.

\bibliographystyle{ACM-Reference-Format}
\bibliography{reference}

\end{document}